\documentclass{article}

\usepackage{times}
\usepackage{graphicx} 
\usepackage{subfig} 

\usepackage{natbib}
\usepackage{algorithm}
\usepackage{algorithmic}
\usepackage{hyperref}

\usepackage{booktabs}
\usepackage{relsize}
\usepackage{lipsum}
\usepackage{tabularx}
\usepackage{amsmath}
\usepackage{amsthm}
\usepackage{amssymb}

\usepackage[accepted]{icml2016} 

\usepackage{verbatim}

\def\eg{{\it e.g.,}}
\def\ie{{\it i.e.,}}

\newcommand{\tablereducebot}{\vspace{-0.02in}}

\icmltitlerunning{Reducing Bias in Production Speech Models}

\begin{document} 

\twocolumn[
\icmltitle{Reducing Bias in Production Speech Models}
\icmlauthor{\mbox{Eric Battenberg}, \mbox{Rewon Child},  \mbox{Adam Coates}, \mbox{Christopher Fougner},  \mbox{Yashesh Gaur}, \mbox{Jiaji Huang}, \mbox{Heewoo Jun}, \mbox{Ajay Kannan}, \mbox{Markus Kliegl}, \mbox{Atul Kumar},  \mbox{Hairong Liu}, \mbox{Vinay Rao},  \mbox{Sanjeev Satheesh}, \mbox{David Seetapun}, \mbox{Anuroop Sriram}, \mbox{Zhenyao Zhu}} {svail@baidu.com}

\icmladdress{Baidu Silicon Valley AI Lab, 
             1195 Bordeaux Avenue, 
             Sunnyvale CA 94089}

\icmlkeywords{machine learning, automatic speech recognition, deep learning}
\vskip 0.2in 
]

\begin{abstract}
    Replacing hand-engineered pipelines with end-to-end deep learning systems has enabled strong results in applications like speech and object recognition. 
    However, the causality and latency constraints of production systems put end-to-end speech models back into the underfitting regime and expose biases in the model that we show cannot be overcome by ``scaling up'', \ie{} training bigger models on more data. In this work we systematically identify and address sources of bias, reducing error rates by up to 20\% while remaining practical for deployment. We achieve this by utilizing improved neural architectures for streaming inference, solving optimization issues, and employing strategies that increase audio and label modelling versatility.
\end{abstract} 

\section{Introduction}
\label{sec:intro}

Deep learning has helped speech systems attain very strong results on speech recognition tasks for multiple languages \cite{xiong2016achieving, amodei2015deep}. One could say therefore that the automatic speech recognition (ASR) task may be considered `solved' for any domain where there is enough training data. However, production requirements such as supporting streaming inference bring in constraints that dramatically degrade the performance of such models -- typically because models trained under these constraints are in the underfitting regime and can no longer fit the training data as well. Underfitting is the first symptom of a model with high bias. In this work, we aim to build a deployable model architecture with low bias because 1) It allows us to serve the very best speech models and 2) Identify better architectures to improve generalization performance, by adding more data and parameters. 

Typically, bias is induced by the assumptions made in hand engineered features or workflows, by using surrogate loss functions (or assumptions they make) that are different from the final metric, or maybe even implicit in the layers used in the model. Sometimes, optimization issues may also prevent the model from fitting the training data as well -- this effect is difficult to distinguish from underfitting, and we also look at approaches to resolve optimization issues.

\textbf{Sources of Bias in Production Speech Models}
   
End-to-end models like \cite{amodei2015deep} typically tend to have lower bias because they have fewer hand engineered features, so we start from a similar model as the baseline. The model used in \cite{amodei2015deep} is a recurrent neural network with two 2D-convolutional input layers, followed by multiple bidirectional recurrent layers and one fully connected layer before a softmax layer. The network is trained end-to-end using the Connectionist Temporal Classification (CTC) loss function \cite{graves2006connectionist}, to directly predict sequences of characters from log spectrograms of the audio. The following assumptions are implicit, that contribute to the bias of the model.

\begin{figure}[t]
\centering
\subfloat[]{\label{fig:ll0}\includegraphics[width=0.24\textwidth]{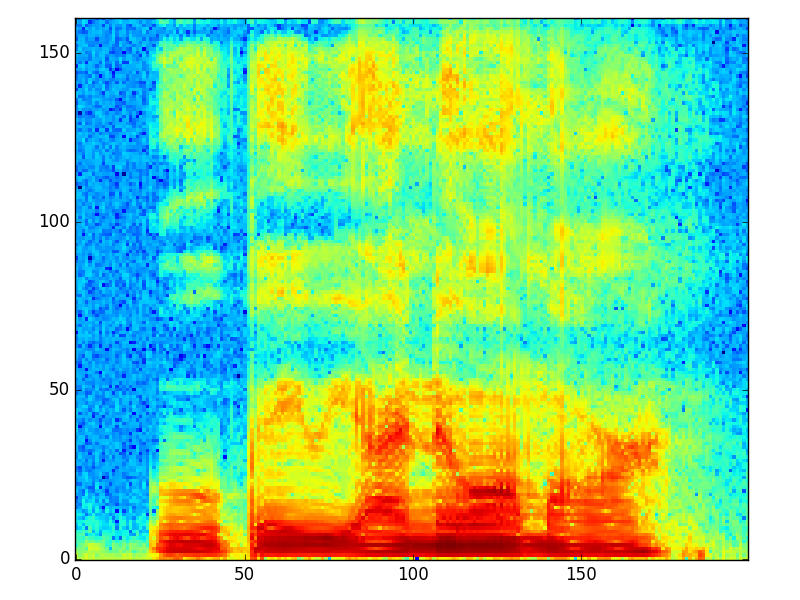}}~
\subfloat[]{\label{fig:lp0}\includegraphics[width=0.24\textwidth]{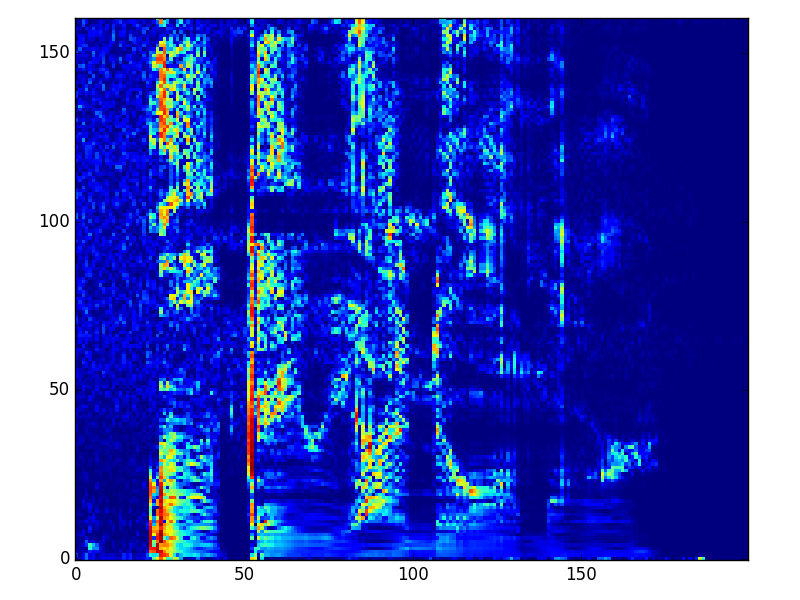}}\\
\subfloat[]{\label{fig:ll2}\includegraphics[width=0.24\textwidth]{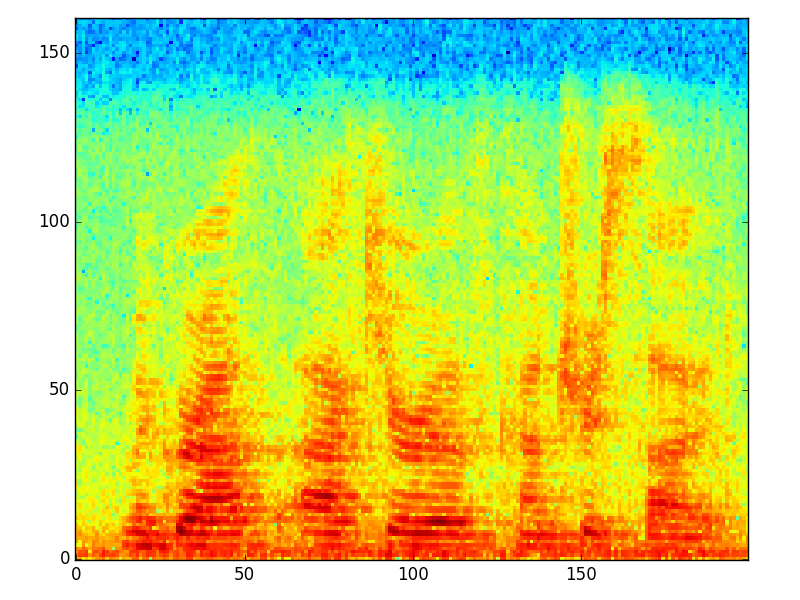}}~
\subfloat[]{\label{fig:lp2}\includegraphics[width=0.24\textwidth]{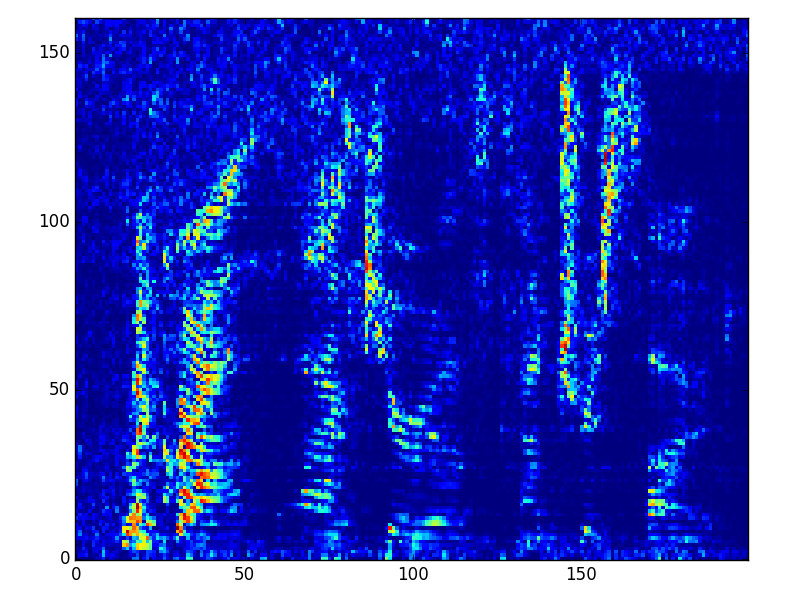}}~
\caption{Each row shows spectrograms of the same audio segment post-processed with two different methods. The horizontal axis is time (10ms / step) and the vertical axis is frequency bins. The left column is generated by applying log. The right is with PCEN with 0.015 and 0.08 smoothing coefficients. In addition, we also want our models to be robust to pipeline effects, like missing bands (bottom row).}
\label{fig:specgrams}
\end{figure}

\begin{enumerate}
	\item {\it Input modeling:}~
	Typically, incoming audio is processed using energy normalization, spectrogram featurization, log compression, and finally, feature-wise mean and variance normalization. Figure~\ref{fig:specgrams} shows however, that log spectrograms can have a high dynamic range across frequency bands (Fig \ref{fig:ll0}) or have some bands missing (Fig \ref{fig:ll2}). We investigate how the PCEN layer ~\cite{wang2016trainable} can parametrize and learn improved versions of these transformations, which simplifies the task of subsequent 2D convolutional layers. 
    \item {\it Architectures for streaming inference:}~English ASR models greatly benefit from using information from a few time frames into the future~\cite{xiong2016achieving, sercu2016dense, peddinti2015time}. In the baseline model, this is enabled by using bidirectional layers, which are impossible to deploy in a streaming fashion, because the backward looking recurrences can be computed only after the entire input is available. Making the recurrences forward-only immediately removes this constraint and makes these models deployable, but also make the assumption that no future context is useful. We show the effectiveness of Latency Constrained Bidirectional RNNs~\cite{Zhang2016HighwayLS} in controlling the latency while still being able to include future context. 
    \item {\it Target modeling:}~CTC models that output characters assume conditional independence between predicted characters given the input features - while this approximation makes maximum likelihood training tractable, this induces a bias on English ASR models and imposes a ceiling on performance. While CTC can easily model commonly co-occuring ngrams together, it is impossible to give roughly equal probability to many possible spellings when transcribing unseen words, because the probability mass has to be distributed between multiple time steps, while assuming conditional independence. We show how GramCTC~\cite{liu2017gramctc} finds the label space where this conditional independence is easier to manage.
    \item {\it Optimization issues:}~Additionally, the CTC loss is notoriously unstable ~\cite{Sak2015}, despite making sequence labeling tractable,  since it is forcing the model to align the input and output sequences, as well as recognize output labels. Making the optimization stable can help learn a better model with the same number of parameters. We show two effective ways of using alignment information to improve the rate of convergence of these models.
\end{enumerate}

The rest of the paper is organized as follows: Section \ref{sec:related_works} introduces related work that address each of the issues outlined above. Sections \ref{sec:pcen}, \ref{sec:lcbrnn}, \ref{sec:gramctc}, and \ref{sec:optimization} investigate solutions for addressing the corresponding issue, and study trade-offs in their application. In section \ref{sec:experiment}, we present experiments where we show the impact of each component independently, as well as the combination of all of them and discuss the results.

\section{Related Work}
\label{sec:related_works}
The most direct way to remove all bias in the input-modeling is probably learning a sufficiently expressive model directly from raw waveforms as in \cite{Sainath2015LearningTS, Zhu2016LearningMF} by parameterizing and learning these transformations. These works suggest that non trivial improvement in accuracy purely from modeling the raw waveform is hard to obtain without a significant increase in the compute and memory requirements. \cite{wang2016trainable} introduced a trainable per-channel energy normalization layer (PCEN) that parametrizes power normalization as well as the compression step, which is typically handled by a static log transform.

Lookahead convolutions have been proposed for streaming inference \cite{chong2016lookahead}. Latency constrained Bidirectional recurrent layers (LC-BRNN) and Context sensitive chunks (CSC) have been proposed in \cite{chen2016training} for tractable sequence model training but not explored for streaming inference. Time delay neural networks \cite{peddinti2015time} and Convolutional networks are also options for controlling the amount of future context.

Alternatives have been proposed to relax the label independence assumption of the CTC loss - Attention models \cite{bahdanau2015, chan2016listen}, global normalization \cite{collobert2016wav2letter} and segmental RNNs \cite{Lu2016SegmentalRN} and more end-to-end losses like lattice free MMI (Maximum Mutual Information) \cite{Povey2016PurelySN} are all promising approaches to address this problem.

CTC model training has been shown to be made more stable by feeding shorter examples first, like SortaGrad \cite{amodei2015deep} and by warm-starting CTC training from a model pre-trained by Cross-Entropy (CE) loss (using alignment information) \cite{Sak2015}. SortaGrad additionally helps to converge to a better training error.

\section{Input modeling}
\label{sec:pcen}


ASR systems often have a vital front-end that involves power normalization, (mel) spectrogram calculation followed by log compression, mean and variance normalization apart from other operations. In this section, we show that we can better model a wide variety of speech input by replacing this workflow with a trainable frontend.

While spectrograms strike an excellent balance between compute and representational quality, they have a high dynamic range (Figure \ref{fig:specgrams}) and are susceptible to channel effects such as room impulse response, Lombard effects and background noises. To alleviate the first issue, they are typically log compressed, and then mean and variance normalized.  However, this only moderately helps with all the variations that can arise in the real world as described before, and we expect the network to learn to be robust to these effects by exposing it to such data. 
By relieving the network of the task of speech and channel normalization, it can devote more of its capacity for the actual speech recognition task. For this, we replaced the traditional log compression and power normalization steps with a \textbf{trainable} per-channel energy normalization (PCEN) front-end \cite{wang2016trainable}, which performs
\begin{equation}
y(t, f) = \left(
  \frac{x(t, f)}{(\epsilon + M(t, f)) ^ {\alpha}} +
  \delta
\right) ^ {r} -
\delta ^ {r},
\label{eq:pcen}
\end{equation}
where $x$ is the input spectrogram, $M$ is the causal energy estimate of the input, and $\delta, \alpha, r, z$ are tunable per-channel parameters. The motivation for this is two-fold. It first normalizes the audio using the automatic gain controller (AGC), $x/M^\alpha$, and further compresses its dynamic range using $(\cdot + \delta)^r - \delta^r$. The latter is designed to approximate an optimized spectral subtraction curve \cite{porter1984optest} which helps to improve robustness to background noises. Clearly, Figure \ref{fig:specgrams} shows that PCEN effectively normalizes various speaker and channel effects.

\begin{figure}[t]
\centering
\subfloat[Homogeneous speech dataset] 
{\label{fig:pcen-bias-wsj}\includegraphics[width=0.24\textwidth]{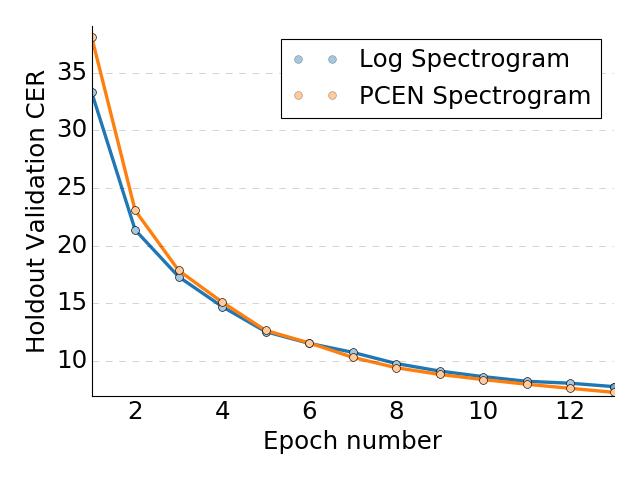}}~
\subfloat[Real world speech dataset] 
{\label{fig:pcen-bias-complete}\includegraphics[width=0.24\textwidth]{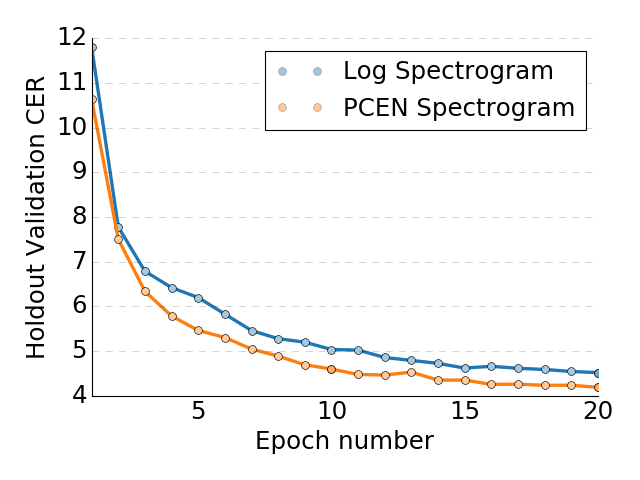}}
\caption{The normalization bias of log compression is clearly shown on the inhomogeneous real-world dataset that is distorted with various channel and acoustical effects.}
\label{fig:pcen-bias}
\end{figure}

PCEN was originally motivated to improve keyword spotting systems, but our experiments show that it helps with general ASR tasks, yielding a noticeable improvement in error rates over the baseline (Table \ref{table:results}). Our training data set which was curated in-house consists of speech data collected in multiple realistic settings. The PCEN front-end gave the most improvement in our far-field validation portion where there was an absolute $\sim$2 WER reduction. To demonstrate that this was indeed reducing bias, we tried this on WSJ, a much smaller and homogeneous dataset. We observed no improvement on the holdout validation set as shown in Figure \ref{fig:pcen-bias-wsj} as the read speech is extremely uniform and the standard front-end suffices.

\section{Latency Controlled Recurrent layers}
\label{sec:lcbrnn}
Consider a typical use-case for ASR systems under deployment. Audio is typically sent over the network in packets of short durations (\eg~ $50$-$200$ ms). Under these streaming conditions, it is imperative to improve accuracy and reduce the latency perceived by end-users. It's observed that users tend to be most perceptive to the time between when they stop speaking and when the last spoken word presents to them. As a proxy for perceived latency, we measure {\it last-packet-latency}, defined as the time taken to return the transcription to the user after the last audio packet arrived at the server.
\footnote{Real-time-factor (RTF) has also been commonly used to measure the speed of an ASR system, but it is in most cases only loosely correlated with latency. While a $RTF < 1$ is necessary for a streaming system, it's far from sufficient. As one example RTF does not consider the non-uniformity in processing time caused by (stacked) convolutions in neural networks.}

\begin{figure}[h]
    \centering
    \includegraphics[width=0.4\textwidth]{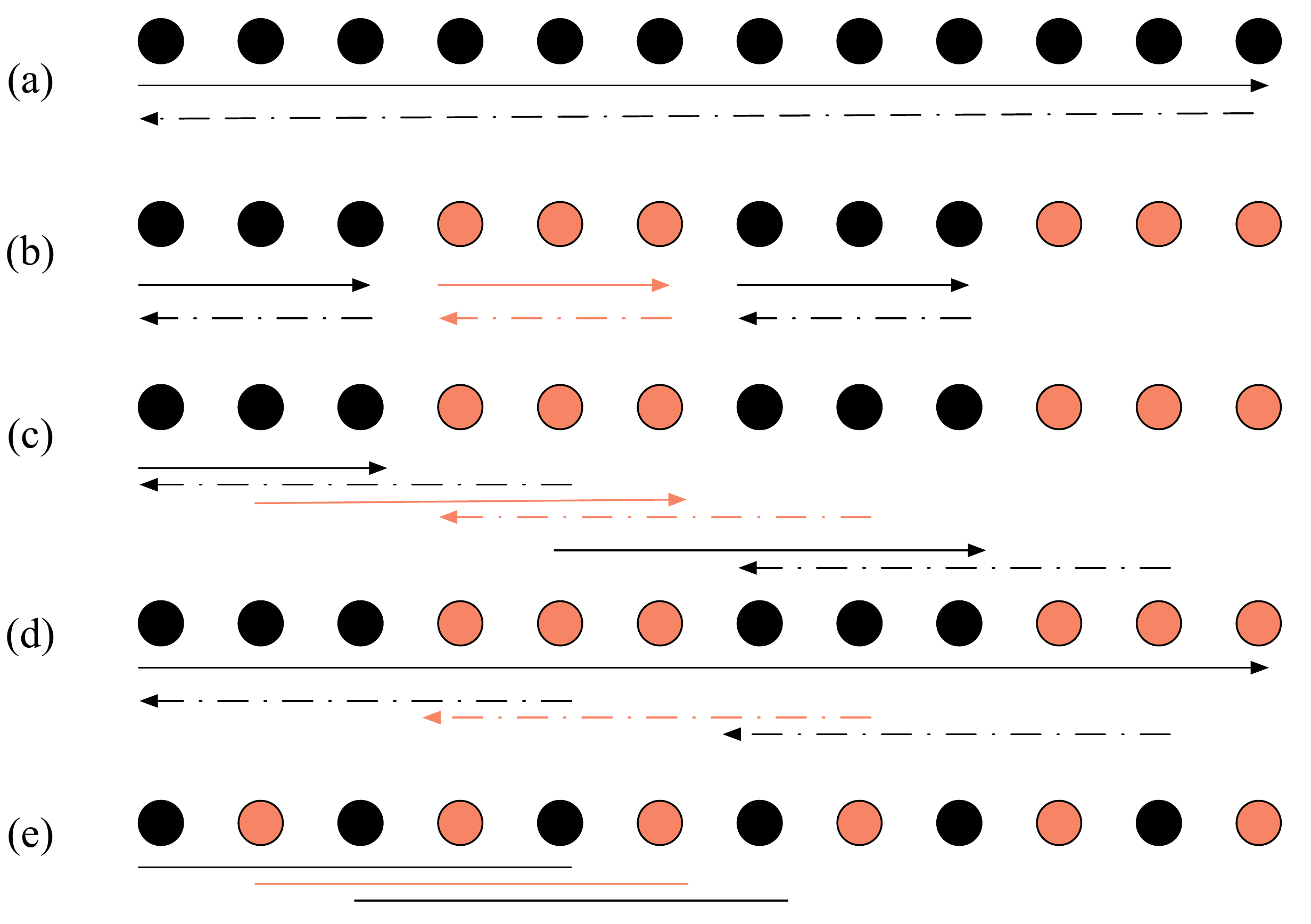}
    \caption{Contexts of different structures. (a) Bidirectional RNN. (b) Chunked RNN. (c) Chunked RNN with overlapping. (d) LC-BGRU layer. (e) Lookahead Convolution.
    Solid arrows represent forward recurrences, and dash arrows represent backward recurrences. States are reset to zero at the start of each arrow. Solid lines represent convolution windows.}
    \label{fig:chunked_rnn}
\end{figure}
 
To tackle the bias induced by using purely forward only recurrences in deployed models, we examine several structures, including look-ahead convolutions \cite{chong2016lookahead} (LA-Conv) and latency-controlled bidirectional RNNs (in our case, LC-BGRU as our recurrent layers employ GRU \cite{gru2014} cells) \cite{chen2016training, Zhang2016HighwayLS}, which are illustrated in Figure \ref{fig:chunked_rnn}.
    
$\bullet$ An LA-Conv layer learns a linear weighted combination (convolution) of activations in the future ([$t$+$1$, $t$+$C$]) to compute activations for each neuron $t$,  with a context size $C$, as shown in Figure \ref{fig:chunked_rnn} (e). The LA-Conv is placed above all recurrent layers.

%
$\bullet$ In a LC-BGRU layer, an utterance is uniformly divided into several overlapping chunks, each of which can be treated as an independent utterance and computed with bidirectional recurrences. More formally, let $L$ be the length of an utterance $X$. $x_i$ represents the $i$th frame of $X$. $X$ is divided into overlapping chunks that are each of a fixed context size $c_W$. In our experiments, the forward recurrences process $X$ sequentially as $x_1$,...,$x_L$. Backward recurrences start processing the first chunk $x_1$,...,$x_{c_W}$, then move ahead by chunk/step-size $c_S$ to independently process $x_{c_S}$,...,$x_{c_W + c_S}$, and so on. In relation to the first chunk $x_1$,...,$x_{c_W}$, we refer to $x_{c_S}$,...,$x_{c_W}$ as the $lookahead$. Hidden-states $h_b$ of the backward recurrences are reset between each chunk, and consequently $h_{b_1}$,...,$h_{b_{c_S}}$ produced from each chunk are used in calculating the final output of the LC-BGRU layer. Figure [\ref{fig:chunked_rnn}] illustrates this operation and compares it with other methods which are proposed for similar purposes. The forward-looking and backward-looking units in this LC-BGRU layer receive the same affine transformations of inputs. We found that this helps reduce computation and save parameters, without affecting the accuracy adversely. The outputs are then concatenated across features at each timestep before being fed into the next layer.

\begin{figure}[t]
    \centering
    \subfloat[]{\label{fig:chunked_cer}\includegraphics[width=0.5\columnwidth]{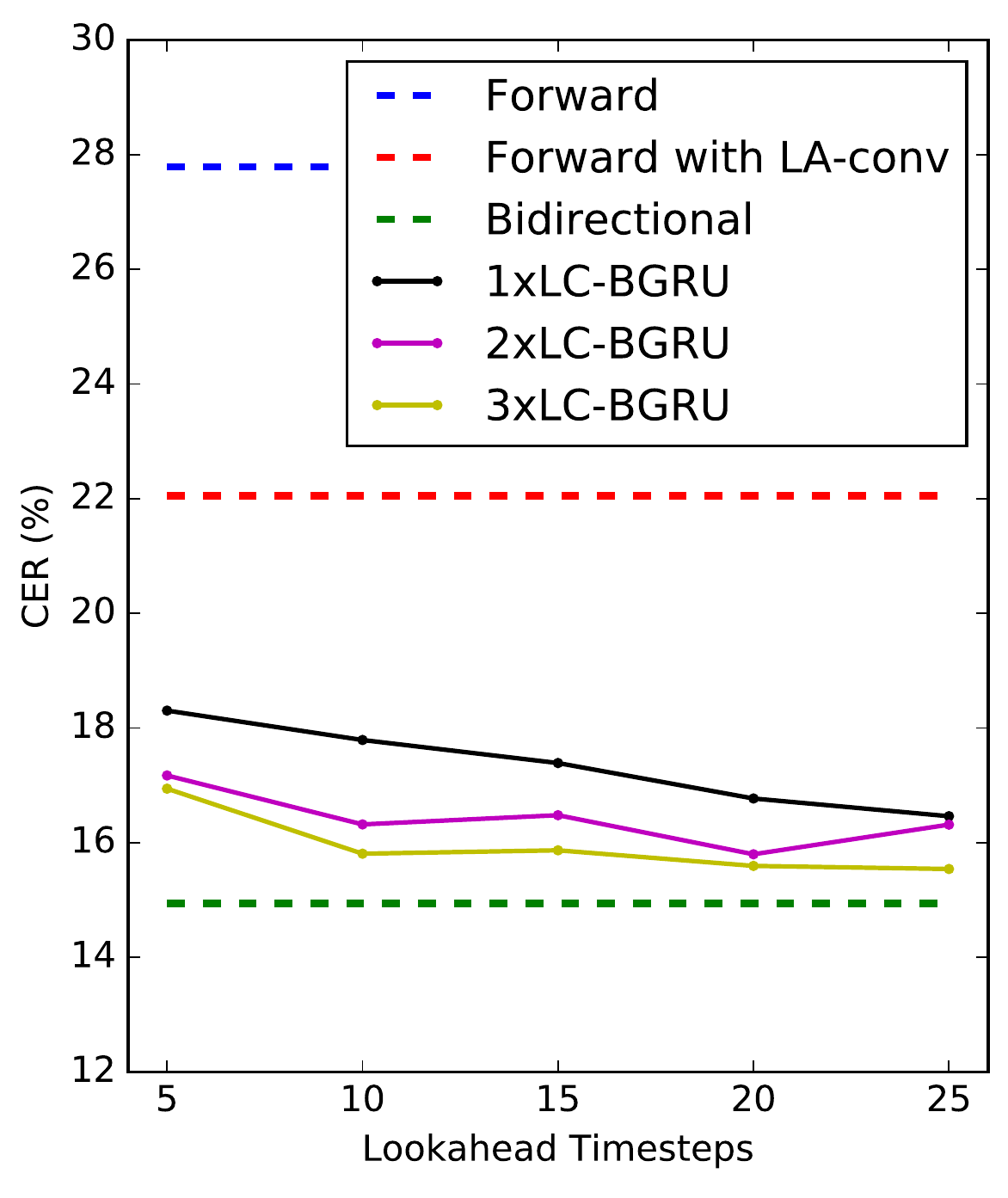}}
    \subfloat[]{\label{fig:chunked_latency}\includegraphics[width=0.5\columnwidth]{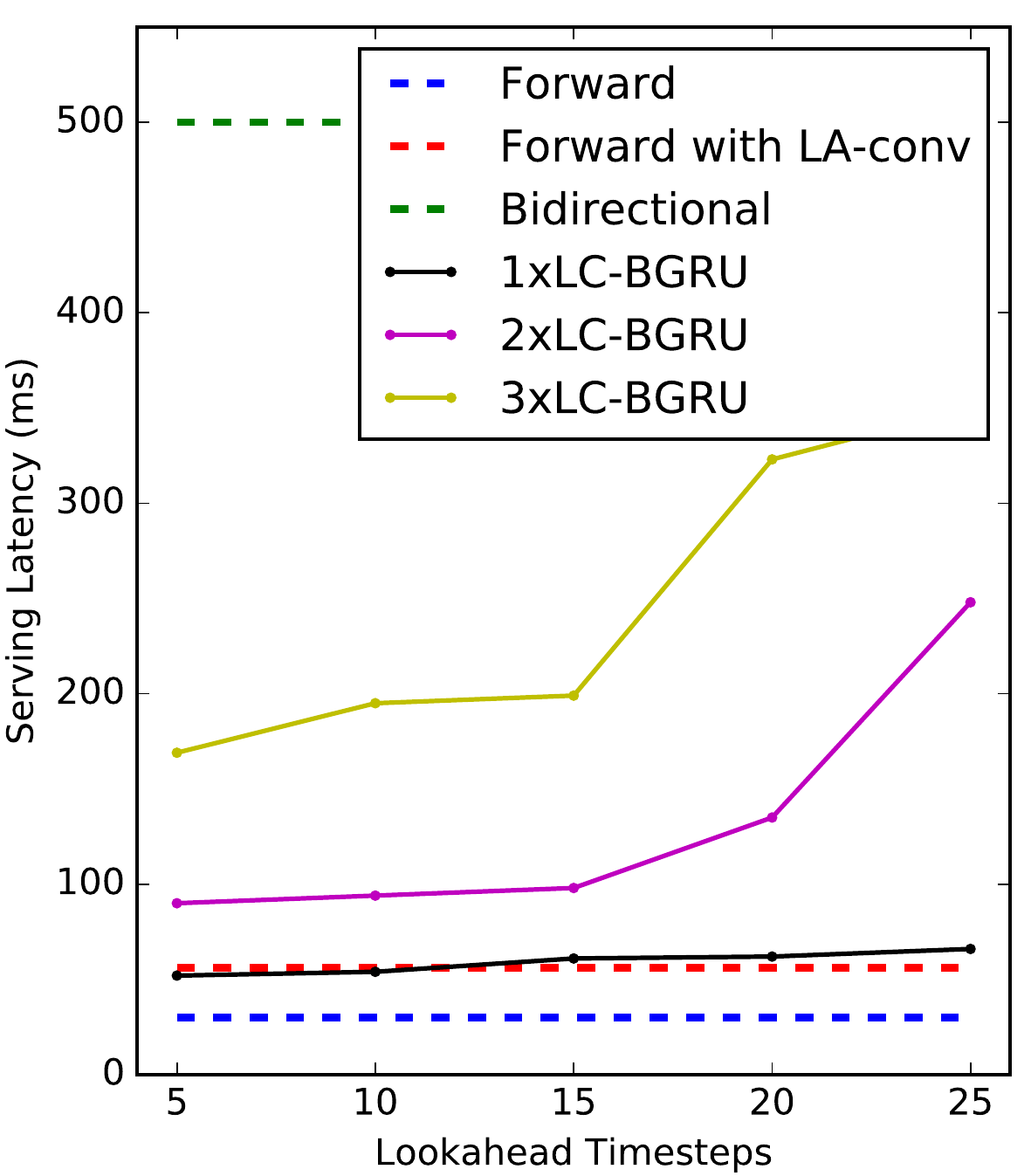}}
    \caption{\ref{fig:chunked_cer} and \ref{fig:chunked_latency} plot CER and serving latency respectively on a dev-set as a function of lookahead, and the number of LC-BGRU layers. The models are trained on a sampled subset (10 $\%$) of the complete training data. The green and blue baselines are the performance of models where all the 3 GRU layers are bidirectional and forward-only correspondingly.}
    \label{fig:latency}
    \vspace{-0.33in}
\end{figure}

\subsection{Accuracy and Serving Latency}
We compare the Character Error Rate (CER) and last-packet-latency of using LA-Conv and LC-BGRU, along with those of forward-GRU and Bidrectional GRU for references. Context size is fixed as 30 time steps for both LA-Conv and LC-BGRU, and lookahead timestep ranges from 5 to 25 every 5 steps for LC-BGRU. For latency experiments, we fix the packet size at $100$ ms, and send one packet every $100$ ms from the client. We send 10 simultaneous streams to simulate a system under moderate load. 
As shown in Figure~\ref{fig:chunked_cer}, while LA-Conv reduces almost half of the gap between forward GRU and bidirectional GRU, a model with three LC-BGRUs with lookahead of 25 each (yellow line) performs as well as bidirectional GRU (green line). The accuracy improves, but the serving latency increases exponentially as we stack LC-BGRU layers, because this increases the effective context much like in convolutional layers. Taking both accuracy and serving-latency into consideration, our final models use 1 LC-BGRU layer, with a lookahead of 20 timesteps (400ms) and step-size of 10 timesteps (200ms). \footnote{1 timestep corresponds to 10ms of the raw-input spectrogram, and then striding in the convolution layers makes that 20ms}

\subsection{Loading BGRU as LC-BGRU}
\begin{figure}[t]
    \centering
    \includegraphics[width=0.65\columnwidth]{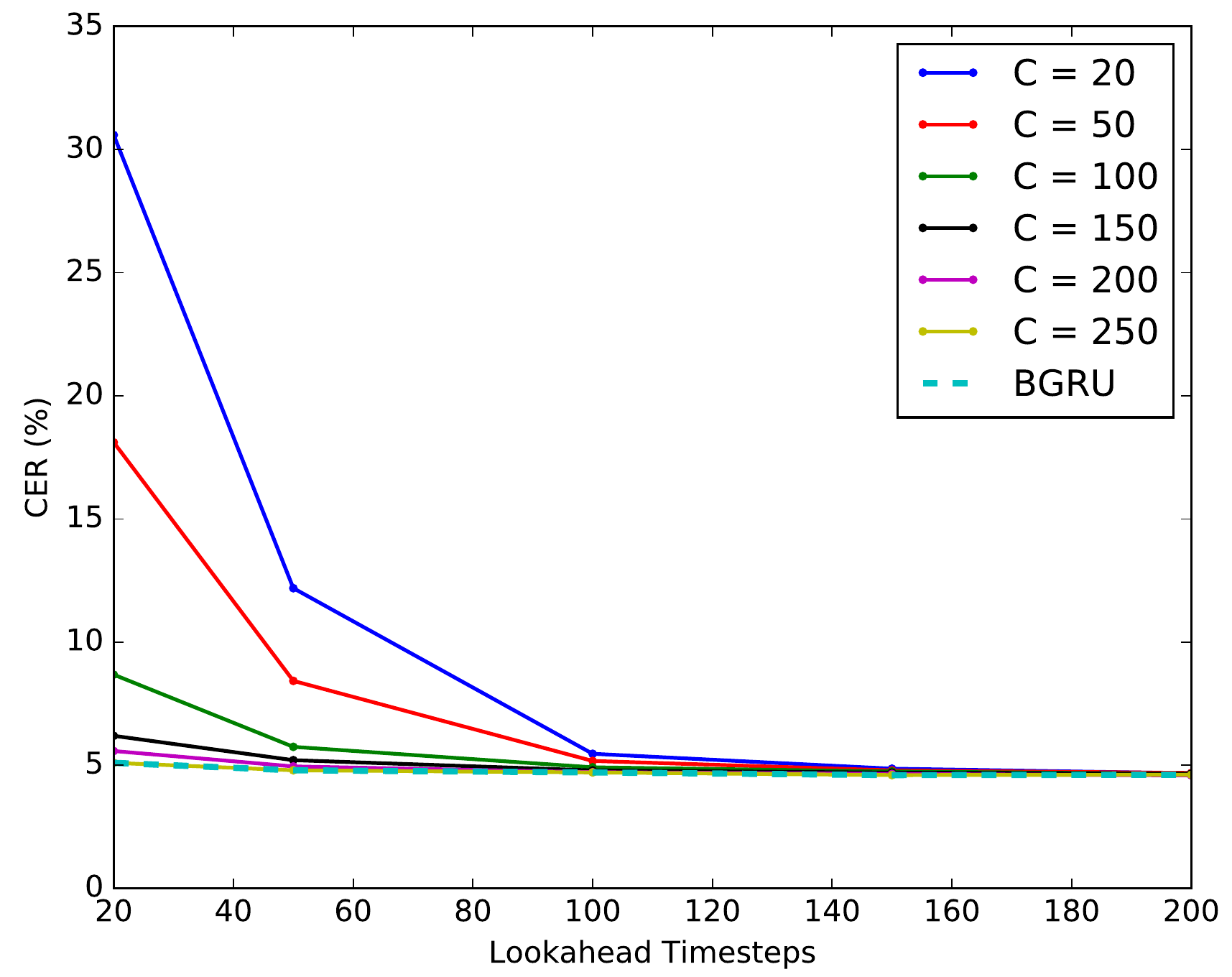}
    \caption{This plots CER of running BGRU in a LC-BGRU way with different context sizes (C) and lookahead timsteps.}
    \label{fig:loading}
    \vspace{-0.3in}
\end{figure}
Since Bidrectional GRUs (BGRU) can be considered as an extreme case of LC-BGRUs with infinite context (as long as the utterance length), it is interesting to know whether we could load a trained bidirectional GRU model as an LC-BGRU, so that we don't have to train LC-BGRUs from scratch. However, we found that loading a model with 3 stacked bidirectional GRUs as stacked LC-BGRUs resulted in significant degradation in performance compared to both the bidirectional baseline and a model trained with stacked LC-BGRUs across a large set of chunk sizes and lookaheads.

We can improve the performance of the model, if we instead chop up the input at each layer to a fixed size $c_W$, such that it is smaller than the effective context. We run an LC-BGRU layer on an input of length $c_W$, then stride the input by $c_S$, discard the last ($c_W$ - $c_S$) outputs, and re-run the layer over the strided input. Between each iteration the forward recurrent states are copied over, but the backward recurrent states are reset each time. The effect of using various $c_W$ and $c_S$ is shown in Figure \ref{fig:loading}. This approach is much more successful in that with $c_W >= 300$ timesteps and $c_S >= 150$ timesteps, we are able to obtain nearly identical error rates to the Bidirectional GRU. With this selection of $c_W$ and $c_S$, the network does twice as much computation as would otherwise be needed, and it also has latencies that are unacceptable for streaming applications. However, it does have the advantage of running bi-directional recurrent layers over arbitrarily long utterances in a production environment at close to no loss in accuracy.

\section{Loss function}
\label{sec:gramctc}

\begin{figure}[t]
    \centering
    \includegraphics[width=0.45\textwidth]{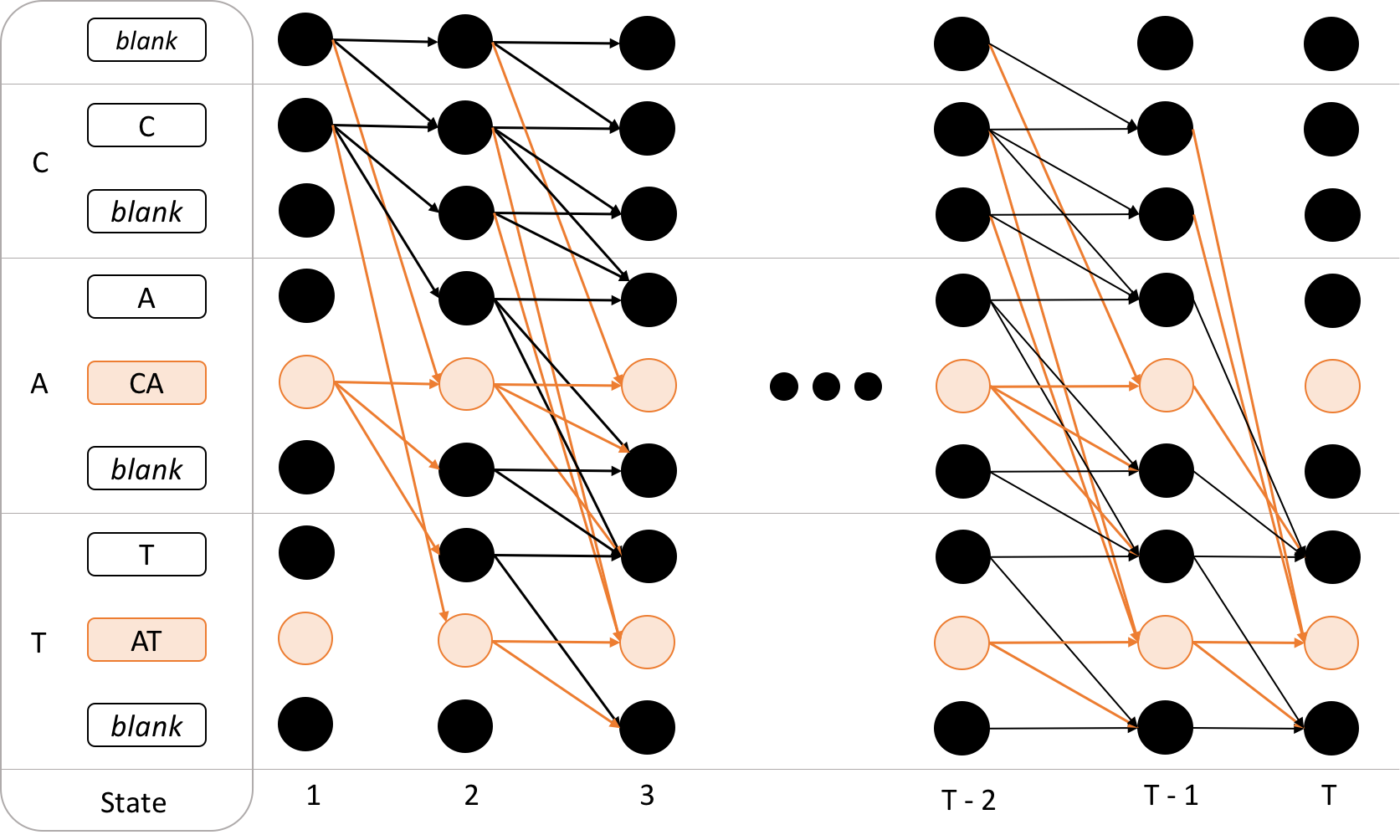}
    \caption{Illustration of the states and the forward-backward transitions for the label `CAT'. Here we let model's output be over the set $G$, the set of all uni-grams and bi-grams of the English alphabet. The set of all valid states for the label $l$ = `CAT' are listed to the left. The set of states and transitions that are common to both CTC and GramCTC are in black, and those that are unique to GramCTC are in orange.}
\label{fig:grams_dp}
\end{figure}

The conditional independence assumption made by CTC forces the model to learn unimodal distributions over predicted label sequences. \textit{GramCTC}~\cite{liu2017gramctc} attempts to find a transformation of the output space where the conditional independence assumption made by CTC is less harmful. Specifically, GramCTC attempts to predict word-pieces, whereas traditional CTC based end-to-end models aim to predict characters. 

GramCTC learns to align and decompose target sequences into word-pieces, or n-grams. N-Grams allow us to address the peculiarities of English spelling and pronunciation, where word-pieces have a consistent pronunciation, but characters don't. For example, when the model is unsure how to spell a sound, it can choose to distribute probability mass roughly equally between all valid spellings of the sound, and let the language model decide the most appropriate way to spell the word. This is often the safest solution, since language models are typically trained on significantly larger datasets and see even the rarest words. GramCTC is a drop-in replacement for the CTC loss function, with the only requirement being a pre-specified set of n-grams $G$. In our experiments, we include all uni-grams and high-frequency bi-grams and tri-grams, which composes a set of 1200 n-grams. 

\subsection{Forward-backward Process of GramCTC}
The training process of GramCTC is very similar to CTC. The main difference is that multiple consecutive characters may form a valid gram. Thus, the total number of states in the forward-backward process is much larger, as well as the transition between these states.

Figure \ref{fig:grams_dp} illustrates partially the dynamic programming process for the target sequence `CAT'. Here we suppose $G$ contains all possible uni-grams and bi-grams. Thus, for each character in `CAT', there are three possible states associated with it: $1$) the current character, $2$) the bi-gram ending in current character, and $3)$ the \emph{blank} after current character. There is also one \emph{blank} at beginning. In total we have $10$ states.

\begin{table}[t]
\begin{center}
\small{
\begin{tabular}{l | c c | c c | c  c}
\toprule
Loss &\multicolumn{2}{c|}{Train} & \multicolumn{2}{c|}{Train\ Holdout} & \multicolumn{2}{c}{Dev} \\
& CER & WER & CER &  WER &  CER &  WER \\
\midrule
 CTC & 4.38 & 12.41 &  4.60 & 12.89 & 11.64 & 28.68\\
 GramCTC & 4.33 & 10.42 &  4.66 & 11.37 & 12.03 & 27.1 \\
\bottomrule
\end{tabular}
}
\end{center}
\vspace{-10pt}
\caption{ Comparison of CTC and GramCTC.}
\label{table:ctcvsgram}
\tablereducebot
\end{table}

\begin{table}[t]
\begin{center}
\begin{tabular}{l | c c | c  c}
\toprule
Loss & \multicolumn{2}{c|}{WER} & \multicolumn{2}{c}{\small{Epoch Time (hours)}}  \\
Stride & 2 &  4 &  2 &  4 \\
\midrule
GramCTC & 21.46 & 18.27 & ~~18.3~ & ~~9.6~\\
\bottomrule
\end{tabular}
\end{center}
\vspace{-10pt}
\caption{Performances and training efficiency of GramCTC with different model strides}
\tablereducebot
\label{table:stride}
\end{table}

\begin{figure*}[h!]
    \centering
    \subfloat[]{\label{fig:delays}\includegraphics[width=0.25\textwidth]{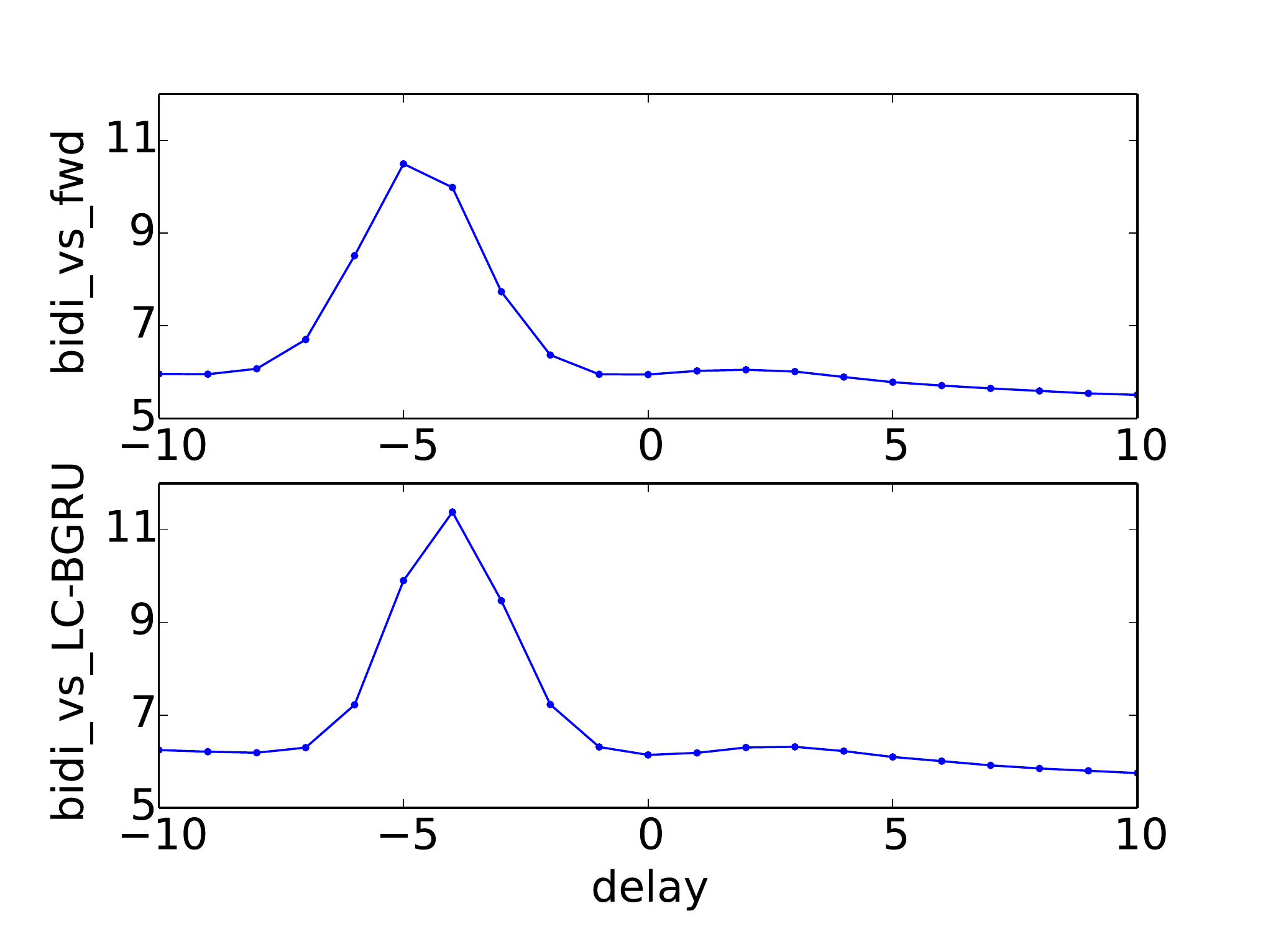}}
    \subfloat[]{\label{fig:delays_effect}\includegraphics[width=0.25\textwidth]{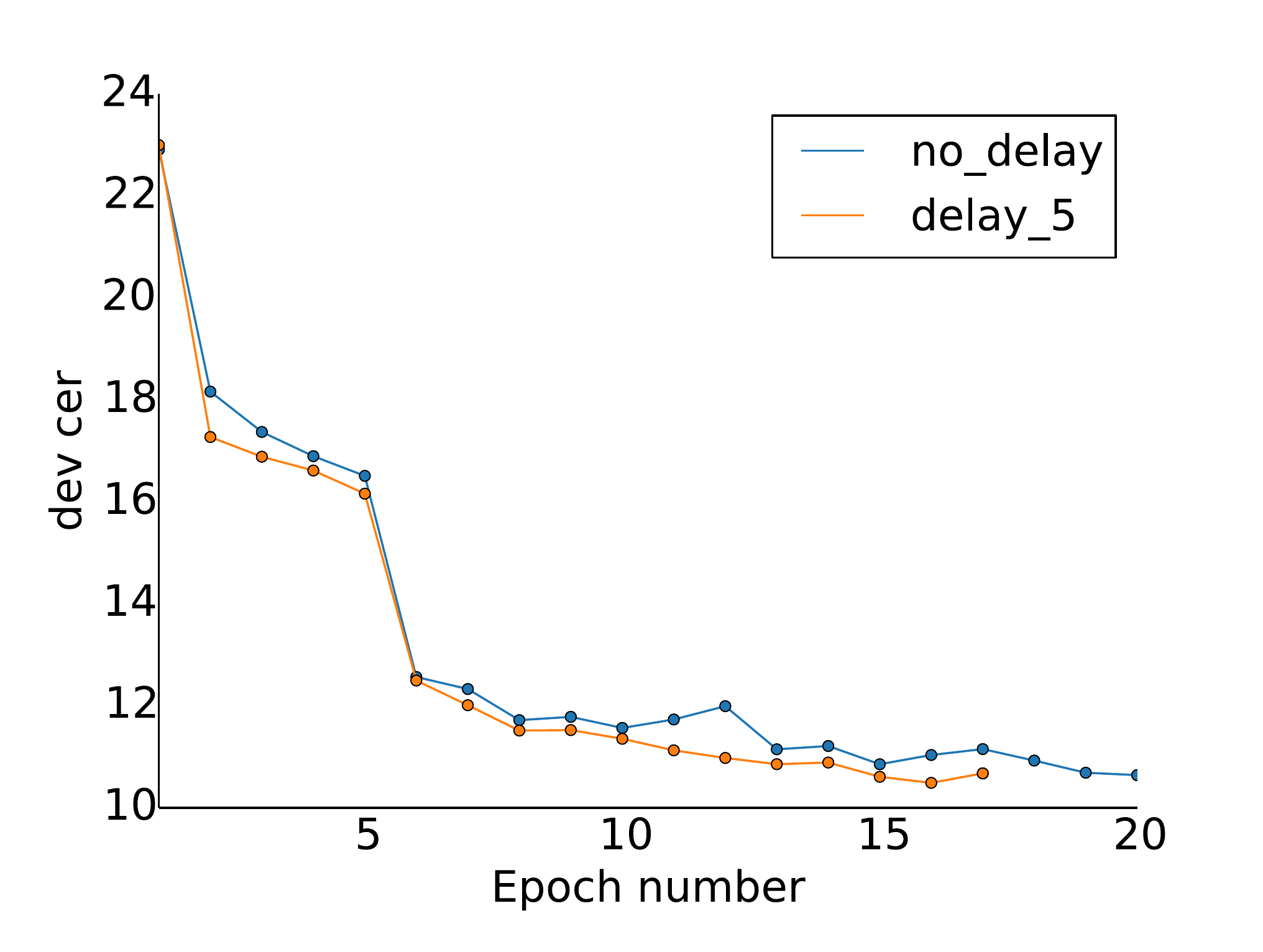}}
    \subfloat[]{\label{fig:alignments_effect_train}\includegraphics[width=0.25\textwidth]{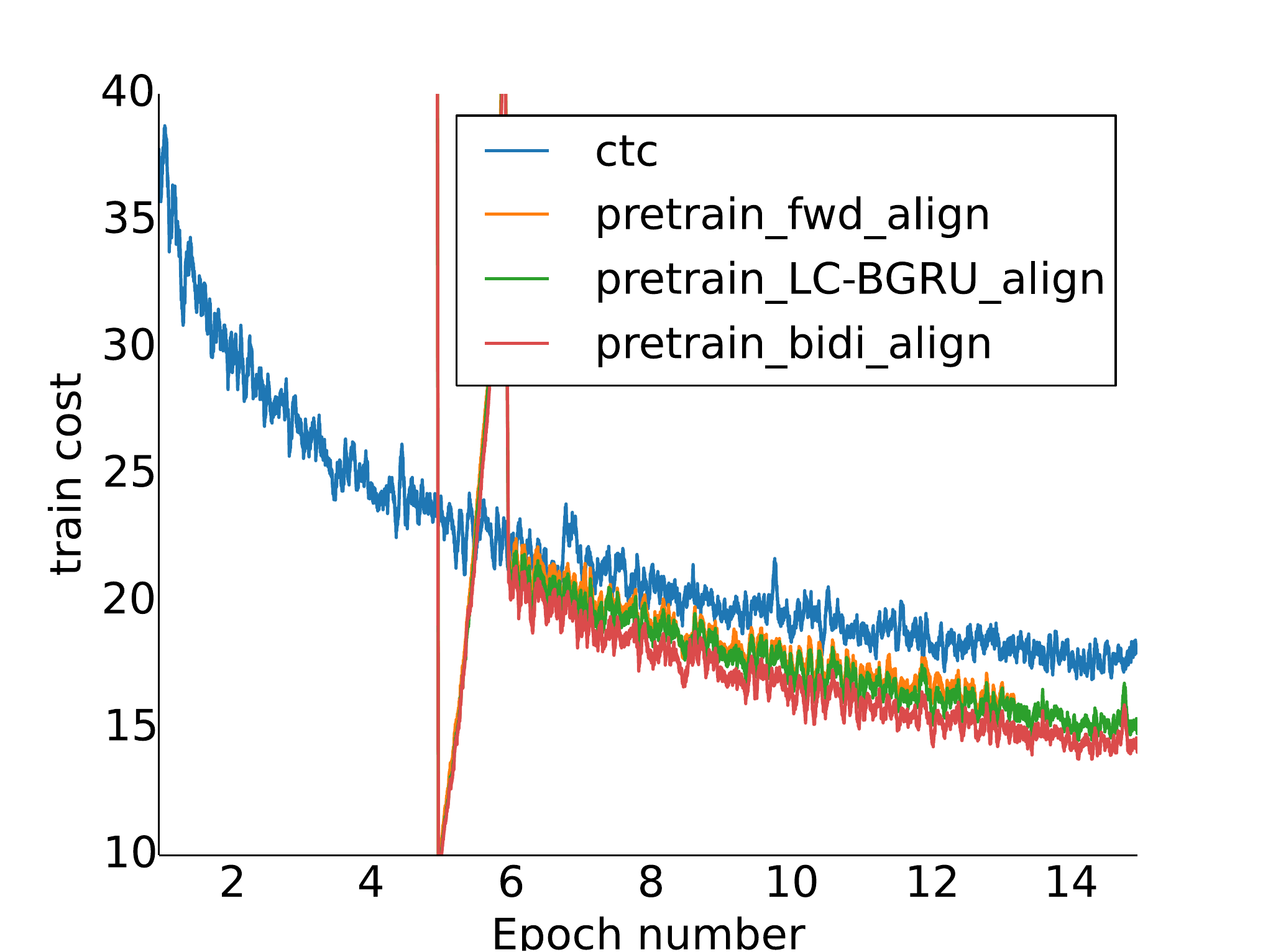}}
    \subfloat[]{\label{fig:alignments_effect}\includegraphics[width=0.25\textwidth]{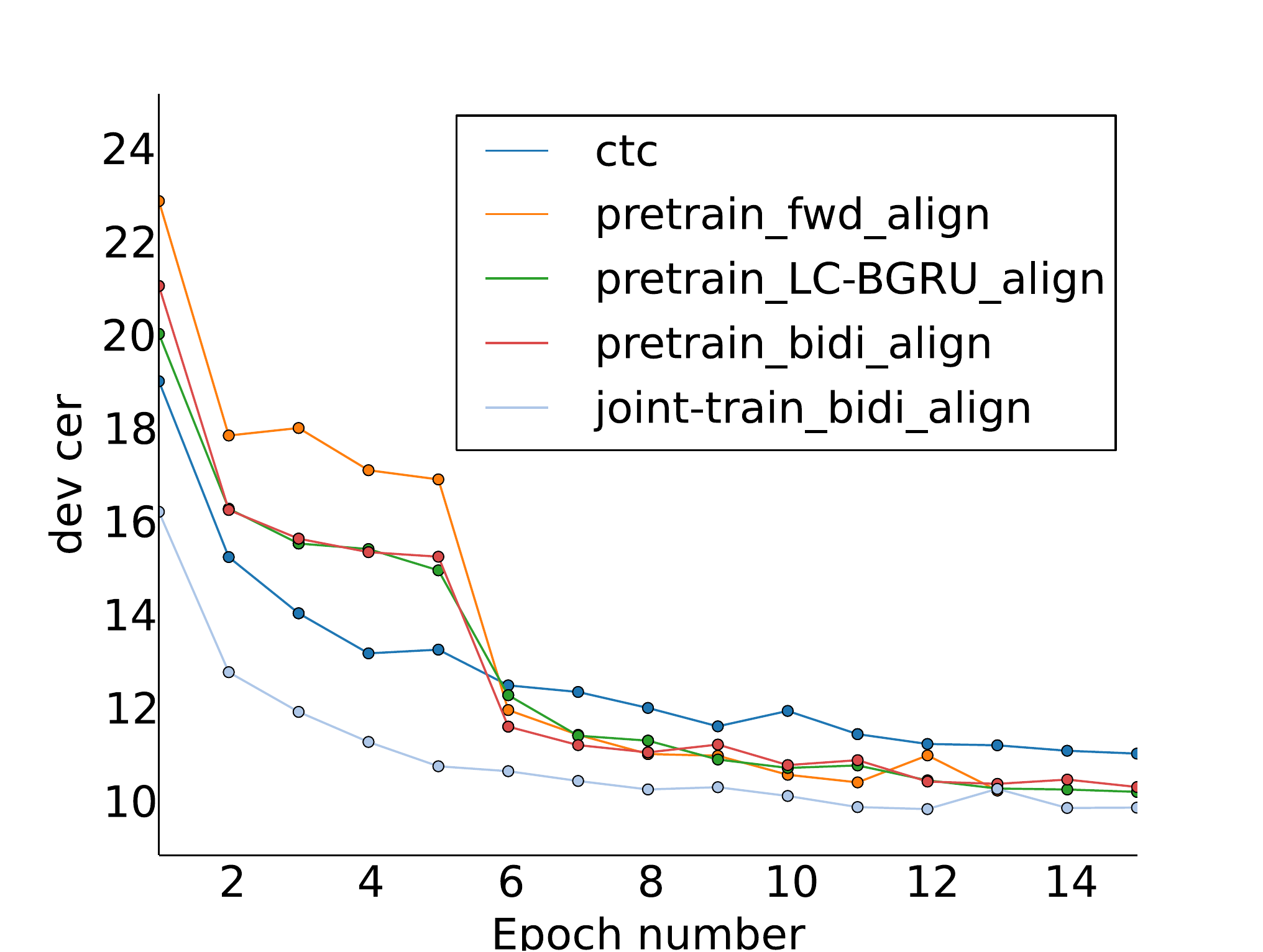}}
    \caption{(a) Cross correlation between alignments estimated by three reference models: forward, LC-BGRU and bidirectional. Alignments by a
    forward (or LC-BGRU) model are 5 (or 4) steps later than those by a         bidirectional model. (b) Applying alignments from a bidirectional model to the pre-training of a forward model, the amount of delay has little impact on the performance at convergence.
    (c) Warm-starting a LC-BGRU model from pre-training using different alignments, all of which achieve smaller training loss than no pre-training.
    (d) CER on dev set for the models trained in (c): they are all smaller than the case of no pre-training. Note also that joint training is on-par with pre-training.}
\end{figure*}

\subsection{GramCTC vs CTC}    
GramCTC effectively reduces the learning burden of ASR network in two ways: $1)$ it decomposes sentences into pronunciation-meaningful n-grams, and $2)$ it effectively reduces the number of output time steps. Both aspects simplify the rules the network needs to learn, thus reducing the required network capacity of the ASR task. Table \ref{table:ctcvsgram} compares the performances between CTC and GramCTC using the same network. There are some interesting distinctions. First, the CERs of GramCTC are similar or even worse than CTC; however, the WERs of GramCTC are always significantly better than CTC. This is probably because GramCTC predicts in chunks of characters and the characters in the same chunk are dependent, thus more robust. 
Secondly, we also observe the performance on the dev set is relatively worse than that on the train holdout. Our dev dataset is not drawn from the same distribution of the training data - this exhibits the potential for GramCTC to overfit even a large dataset.

Table \ref{table:stride} compares the training efficiency and the performance of trained model with GramCTC on two time resolutions, $2$ and $4$. By striding over the input at a faster rate in the early layers, we effectively reduce the time steps of later layers, and reduce the training time in half. From stride $2$ to stride $4$, the performance also improves a lot probably because larger n-grams align with larger segments of utterance, and thus need lower time resolution.

\section{Optimization Tricks}
\label{sec:optimization}

Removing optimization issues have been a reliable way of improving performance in deep neural networks~\cite{ioffe2015, He2016DeepRL}. Several optimization tricks have been proposed especially for training recurrent networks - we tried using LayerNorm~\cite{Ba:2016LayerNorm}, Recurrent batch norm~\cite{CooijmansBLC16RecurBN} and NormProp~\cite{Arpit2016NormProp} without much success. Additionally, we take care special care to optimize layers properly, and also employ SortaGrad~\cite{amodei2015deep}.

~\cite{Sak2015} suggests that CTC training could be suffering from optimization issues and could be made more stable by providing alignment information during training. In this section we study how alignment information can be used effectively.

\subsection{Pre-training vs Joint-training}
Using alignment information for training CTC models appears counter intuitive since CTC marginalizes over all alignments during training. However, the CTC loss is hard to optimize because it simultaneously estimates network parameters and alignments. To simplify the problem, one may propose an Expectation-Maximization (EM) like approach, where the E-step computes the expected log-likelihood by marginalizing over the posterior of alignments, and the M-step refines the model parameters by maximizing the expected log-likelihood. However, it is infeasible to compute the posterior for all the  alignments, and we approximate it by taking only the most probable alignment. One step of EM can be considered as the pre-training approach of using alignment information - we start training a model with the most likely alignment (which simplifies to training with a Cross-Entropy (CE) loss for a few epochs, followed by training with the CTC loss. 

Another way of using the alignment information is train a single model simultaneously using a weighted combination of the CTC loss and the CE loss. 

Figure~\ref{fig:alignments_effect_train} shows the training curves of the same model architecture with pure CTC training, pre-training and joint training with alignment information from different source models. In the case of pre-training we stop providing alignment information at the 6-th epoch, corresponding to the shift in the training curve. Note that the final training losses of both pre-trained and joint-trained models are all lower than the pure CTC trained model, showing the effectiveness of this optimization trick. Additionally, joint-training and pre-training are on par in terms of training, so we prefer joint-training to avoid multi phase training. The corresponding CER on dev set is presented in figure~\ref{fig:alignments_effect}.

\subsection{Source of alignments}
It is important for us to understand how accurate the alignment information needs to be, since different models have differing alignments according to the architecture and training methods. 

We estimate alignments from three ``reference" models (models with forward only GRU, LC-BGRU and bidirectional GRU layers, all trained with several epochs of CTC minimization), and present the cross correlation between the alignments produced by these models in Fig.~\ref{fig:delays}. The location of the peak implies the amount of delays between two alignments. It is evident that alignments by a forward (and LC-BGRU) model are 5 (4) time-steps later than those by a bidirectional model, an observation that is consistent with \cite{schuster1997bidirectional}. Therefore, it seems important to pre-train a model with properly adjusted alignments, e.g., alignments from a bidirectional model are supposed to be delayed for 5 steps to be used in the pre-training of a forward model. However, we found that for models trained on large datasets, this delay has little impact on the final result (figure~\ref{fig:delays_effect}). To push this series of experiments to the extreme, we tried pre-training a model with random alignments. Random alignments do not work, but we found that most likely alignment as predict by any ctc model was sufficient to achieve improved optimization.

\section{Experiments}
\label{sec:experiment}

\subsection{Setup}
In all experiments, the dataset is 10,000 hours of labeled speech from a wide variety of sources. The dataset is expanded by noise augmentation -- in every epoch, $40\%$ of the utterances are randomly selected and background noise is added. For robustness to reverberant noise encountered in far-field recognition, we adopt room impulse response (RIR) augmentation as in \cite{ko2017reverberant}, in which case, we randomly sample a subset of the data and convolve each instance with a random RIR signal. \footnote{We collect RIRs by emitting a signal from a speaker and capturing the signal, as well as the reverberations from the room, using an linear array of 8 microphones. The speaker is placed in a variety of configurations, ranging from 1 to 3 meters distance and 60 to 120 degrees inclination with respect to the array, for 20 different rooms.} 

\begin{figure}[t]
    \centering
    \includegraphics[width=1\linewidth]{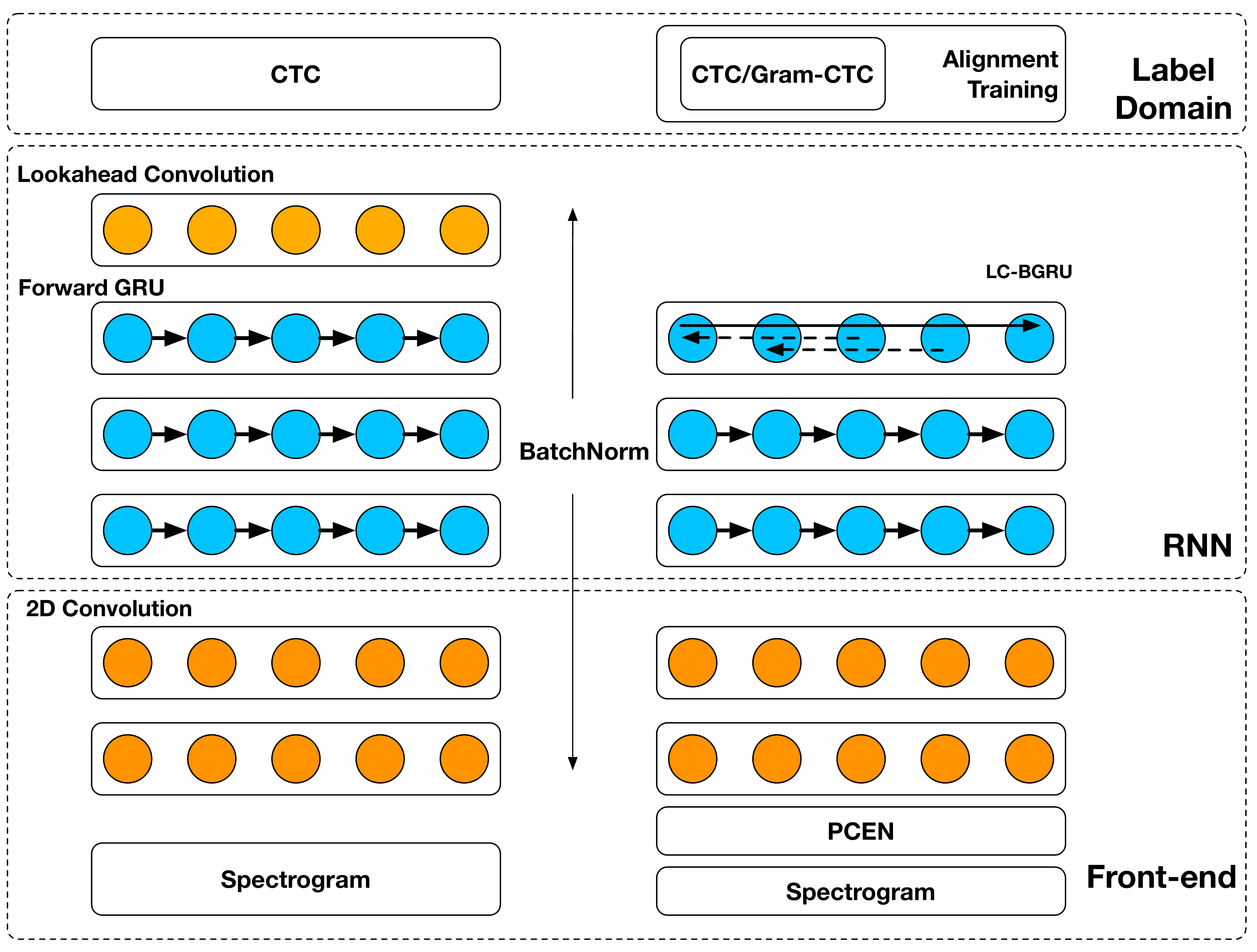}
    \caption{Comparison between the baseline ({\it left}) and the proposed ({\it right}) model architectures.}
    \label{fig:ds21}
\end{figure}

The model specification and training procedure are the same as in \cite{amodei2015deep}. The baseline model is a deep recurrent neural network  with two 2D convolutional input layers, followed by 3 forward Gated Recurrent layers \cite{gru2014}, 2560 cells each, a look-ahead convolution layer and one fully connected layer before a softmax layer. The network is trained end-to-end to predict characters using the CTC loss. The configurations of the 2D convolution layers (filters, filter dimensions, channels, stride) are (32, 41x11, 1, 2x2) and (32, 21x11, 32, 2x1). Striding in both time and frequency domains helps us reduce computation in the convolution layers. In the convolution and fully-connected layers, we apply batch-normalization before applying nonlinearities (ReLU). We use sequence-wise batch-normalization in the recurrent layers \cite{amodei2015deep}, effectively acting on the affine transformations of the inputs fed into them. Figure~\ref{fig:ds21} shows the baseline model on the left. 

For the baseline model, log spectrogram features are extracted, in 161 bins with a hop size of 10ms and window size of 20ms, and are normalized so that each input feature has zero mean and unit variance. The optimization method we use is stochastic gradient descent with Nesterov momentum. Hyperparameters (batch-size = $512$, learning-rate $7\times 10^{-4}$, momentum $0.99$) are kept the same across different experiments. 

\label{sec:results}
Table~\ref{table:results} shows the result of the proposed solutions in earlier sections. We report the results on a sample of the train set as well as a development set. The error rates on the train set are useful to identify over-fitting scenarios, especially since the development set is significantly different from our training distribution as their sources are different. 

In Table~\ref{table:results}, both character and word error rates (CER/WER) are produced using a greedy max decoding of the output softmax matrix, i.e., taking the most likely symbol at each time step and then removing blanks and repeated characters.  However, when a language model is adopted as in the ``Dev LM" results, we use a beam search over the combined CTC and LM scores.  

\begin{table*}[ht]
\centering
\begin{tabular}{l|cc|cccc|cc}
\toprule
          & \multicolumn{2}{c|}{Train} & \multicolumn{4}{c|}{Dev} & \multicolumn{2}{c}{Dev LM}  \\
                                 & CER   & WER   & CER   & \% Rel  & WER   & \% Rel  & WER    & \% Rel  \\
\midrule
Baseline                         & 4.38  & 12.41 & 11.64 & 0.00\%  & 28.68 & 0.00\%  & 18.95  & 0.00\%  \\
\midrule
\textbf{Individual-changes} & & & & & & & \\
\midrule
Baseline + PCEN                  & 3.79  & 10.90 & 11.16 & 4.20\%  & 27.85 & 2.90\%  & 18.12  & 4.40\%  \\
Baseline + $1 \times$ LC-BGRU             & 3.49  & 10.33 & 11.06 & 5.00\%  & 27.03 & 5.80\%  & 17.47  & 7.80\%  \\
Baseline + GramCTC               & 4.33  & 10.42 & 12.03 & -3.30\% & 27.10 & 5.50\%  & 19.26  & -1.70\%   \\
Baseline + CE pre-training       & 3.31  & 9.50  & 10.84 & 6.90\%  & 26.39 & 8.00\%  & 17.89  & 5.60\%  \\
Baseline + CE joint-training     & 3.25  & 9.58  & 10.75 & 7.70\%  & 26.64 & 7.10\%  & 17.93  & 5.40\%  \\
Baseline + Farfield augmentation & 5.25  & 14.59 & 11.49 & 1.30\%  & 29.64 & -3.30\% & 18.47  & 2.50\%  \\
\midrule
\textbf{Incremental-changes} & & & & & & & \\
\midrule
Baseline + CE + CTC              &       &       &       &         &       &         &        &         \\
\ \ \ + GramCTC joint training (Mix-1)   & \textbf{2.97}  & \textbf{7.31}  & 10.91 & 6.30\%  & 24.48 & 14.60\% & 17.71  & 6.5\%   \\
Baseline + PCEN                  &       &       &       &         &       &         &        &         \\
\ \ \ +$1 \times$ LC-BGRU                &       &       &       &         &       &         &        &         \\
\ \ \ + Farfield augmentation (Mix-2)   & 5.51  & 14.10 & 9.74  & 16.40\% & 24.82 & 13.40\% & \underline{\textbf{15.47}}  & 18.40\% \\
\ \ \ + CE joint training (Mix-3)        & 3.57  & 10.50 & \textbf{9.38}  & 19.40\% & \textbf{23.77} & 17.10\% & 15.75  & 16.90\% \\
\midrule
Bidirectional target             & 2.58  & 7.47  & 9.37  & 19.60\% & 23.03 & 19.70\% & 15.96  & 15.80\% \\
\bottomrule
\end{tabular}
\caption{Results for both single improvements and incremental improvements to the models.  Except when using a language mode (Dev LM), reported numbers are computed using greedy max decoding as described in \ref{sec:results}. Best results using deployable models are bolded.}
\label{table:results}
\end{table*}

\subsection{Results of Individual Changes}
\label{sec:indiv-changes}
In the first half of Table~\ref{table:results}, we show the impact of each of the changes applied individually. All of the techniques proposed help fit the training data better, measured by CER on the train set. Several observations stand out.
\begin{enumerate}
\item \textbf{Replacing CTC loss with GramCTC loss} achieves a lower WER, while CERs are similar on the train set. This indicates that the loss promotes the model to learn the spelling of words, but completely mis-predicts words when they are not known. This effect results in diminished improvements when the language model is applied.

\item Applying \textbf{farfield augmentation} on the same sized model results in a worse training error as expected. It shows a marginal improvement on the dev set, even though our dev set has a heavy representation of farfield audio. 

\item The single biggest improvement on the dev set is the addition of the \textbf{LC-BGRU} which closes the gap to bidirectional models by 50\%. 

\item \textbf{Joint (and pre) training with alignment information} improves CER on the train set by 25\%, highlighting optimization issues in training CTC models from scratch.  However, these models get less of an improvement from language model decoding, indicating their softmax outputs could be overconfident, therefore less amenable to correction by the language model.  This phenomenon is observed in all models employing CE training as well as our Bidirectional target model (the model that provides the targets used for CE training).  
\end{enumerate}

\subsection{Results of Incremental Changes}
While we designed the solutions to address distinct issues in the model, we should not expect every individual improvement to be beneficial when used in combination. As an example, we see in the section on optimization that models with bidirectional layers gain very little by using alignment information - clearly, bidirectional layers by themselves address a part of the difficulty in optimizing CTC models. Therefore, addressing the absence of bidirectional layers will also address optimization difficulties and they may not stack up.

We see in the second half of Table~\ref{table:results} that improvements indeed do not stack up. There are 3 interesting models to discuss.

\begin{enumerate}
\item The model mix of joint training with 3 increasingly difficult losses (CE, CTC, and GramCTC, Mix-1) achieves the best results on the train set far surpassing the other model mixes, and even nearly matching the performance of models with bidirectional layers on the train set. This model has the smallest gain on the dev set amongst all the mix-models, and puts it in the overfitting regime.  We know that there exists a model that can generalize better than this one, while achieving the same error rates on the train set: the bidirectional baseline. Additionally, this model receives a weak improvement from the language model, which agrees with what we observed with GramCTC and CE training in \ref{sec:indiv-changes}. 

\item The model mix of PCEN, LC-BGRU and IR augmentation (Mix-2) performs worse on the train set -- additional data augmentation with IR impulses makes the training data harder to fit as we have seen earlier, but PCEN and LC-BGRU is not sufficient to address this difficulty. However, the model does attain better generalization and performs better on the dev set, and actually surpasses our bidirectional target when using a language model.

\item Mix-3 adds CE joint training which helps to address optimization issues and leads to lower error rates on both the train and dev sets. However, the improvement in dev WER disappears when using a language model, again highlighting the language model integration issues when using CE training. 

\end{enumerate}
%
%
%
Finally in Table~\ref{table:final}, we compare Mix-3 against the baseline model, and its equivalent with twice as many parameters in every layer, on various categories of speech data. Clearly, Mix-3 is significantly better for ``farfield" and ``Names" speech data, two notably difficult categories for ASR. ASR tasks run into a generalization issue for ``Names" categories because they are often required words that is not present in the acoustic training data. Similarly far field audio is hard to obtain and the models are forced to generalize out of the training data, in this case by making use of augmentation. At the same time, the serving latency of Mix-3 is only slightly higher than baseline model, still good for deployment.

\begin{table}[h]
\centering
\begin{tabular}{l  c  c  c}
\toprule
Devsets & Baseline & 2$\times$Baseline* & Mix-3  \\
\midrule
Clean casual speech & 5.90 & \textbf{5.00} & 5.80 \\
Farfield & 35.05 & 30.60 & \textbf{26.49} \\
Names & 19.73 & 19.30 & \textbf{17.40} \\
Overall & 18.46 & 17.46 & \textbf{15.74} \\
\midrule 
Serving latency & \textbf{112} & 25933 & 153 \\
Training time & \textbf{17} & 29 & 25 \\
\bottomrule
\end{tabular}
\caption{A deeper look into the dev set, measuring WER with language model decoding of each model on different slices of the dev set. Serving latency (milliseconds) is the $98^{th}$ percentile latency on the last packet as described in Sec. \ref{sec:lcbrnn}. Training time is in hours per epoch with the data and infrastructure the same. *This model has twice the number of parameters as the Baseline, and suffers from prohibitively large serving latency.}
\label{table:final}
\end{table}

\section{Conclusion}
In this work, we identify multiple sources of bias in end-to-end speech systems which tend to encourage very large neural network structure, thus make deployment impractical. Multiple methods are proposed to address these issues, which enable us to build a model that performs significantly better on our target dev set, while still being good for streaming inference. 

While the addition of cross entropy alignment training and the GramCTC loss allow models to fit the training and validation data better with respect to the WER of a greedy max decoding, they see much less of a benefit from language modeling integration.  Using an LC-BRGU layer in place of lookahead convolutions conveys benefits across the board as does use of a PCEN layer at the front end.  Finally, generalization to unseen data is improved by the addition of farfield augmentation.
    
\section{Acknowledgements}
We are indebted to the Baidu Speech Technology Group for all the IR convolutions, and the Quality Assurance team for helping us identify and understand useful metrics (like first, last and 98\% packet latency). We are also grateful to the systems team at SVAIL, for their help in developing the training platform and infrastructure.

\bibliography{ds2.1}

\end{document}